\documentclass[conference]{IEEEtran}
\usepackage[compatibility=false]{caption}

\makeatletter

\def\ps@IEEEtitlepagestyle{%
  \def\@oddfoot{\mycopyrightnotice}%
  \def\@evenfoot{}%
}
\def\mycopyrightnotice{%
  {\footnotesize XXX-X-XXXX-XXXX-X/XX/\$XX.00~\copyright~20XX IEEE\hfill}
  \gdef\mycopyrightnotice{}
}

\usepackage{blindtext}
\usepackage{eso-pic}
\IEEEoverridecommandlockouts
\usepackage{cite}
\usepackage{amsmath,amssymb,amsfonts}
\usepackage{algorithm} 
\usepackage{algpseudocode} 
\usepackage{graphicx}
\usepackage{textcomp}
\usepackage{xcolor}
\usepackage[table]{xcolor}  
\usepackage{algpseudocode}
\usepackage{enumerate} 
\usepackage{subcaption}
\def\BibTeX{{\rm B\kern-.05em{\sc i\kern-.025em b}\kern-.08em
    T\kern-.1667em\lower.7ex\hbox{E}\kern-.125emX}}
    
\usepackage{eso-pic}
\newcommand\AtPageUpperMyright[1]{\AtPageUpperLeft{%
 \put(\LenToUnit{0.17\paperwidth},\LenToUnit{-2cm}){%
     \parbox{0.9\textwidth}{\raggedleft\fontsize{8}{11}\selectfont #1}}%
 }}%
\newcommand{\conf}[1]{%
\AddToShipoutPictureBG*{%
\AtPageUpperMyright{#1}
}
}

\begin{document}
\title{\vspace*{1cm} Volatility in Certainty (VC): A Metric for Detecting Adversarial Perturbations During Inference in Neural Network Classifiers\\
\thanks{This research is supported by (the NASA-ULI under Cooperative Agreement Number 80NSSC20M0161. Also, this work partially supported by the University Transportation System (UTC), Department of Transportation, USA, through grant number 69A3552348327.}
}

\author{\IEEEauthorblockN{Vahid Hemmati}
\IEEEauthorblockA{\textit{Electrical and Computer Engineering} \\
\textit{North Carolina A\&T State University}\\
Greensboro, USA \\
Hemmati@ncat.edu}
\and
\IEEEauthorblockN{Ahmad Mohammadi}
\IEEEauthorblockA{\textit{Electrical and Computer Engineering} \\
\textit{North Carolina A\&T State University}\\
Greensboro, USA \\
amohammadi@aggies.ncat.edu}
\and
\IEEEauthorblockN{Abdul-Rauf Nuhu}
\IEEEauthorblockA{\textit{Electrical and Computer Engineering} \\
\textit{North Carolina A\&T State University}\\
Greensboro, USA \\
anuhu@aggies.ncat.edu}
\and
\IEEEauthorblockN{Reza Ahmari}
\IEEEauthorblockA{\textit{Electrical and Computer Engineering} \\
\textit{North Carolina A\&T State University}\\
Greensboro, USA \\
rahmari@aggies.ncat.edu}
\and
\IEEEauthorblockN{Parham Kebria}
\IEEEauthorblockA{\textit{Electrical and Computer Engineering} \\
\textit{North Carolina A\&T State University}\\
Greensboro, USA \\
pmkebria@ncat.edu}
\and
\IEEEauthorblockN{Abdollah Homaifar}
\IEEEauthorblockA{\textit{Electrical and Computer Engineering} \\
\textit{North Carolina A\&T State University}\\
Greensboro, USA \\
homaifar@ncat.edu}
}

\maketitle
\conf{\textit{ 5. Interdisciplinary Conference on Electrics and Computer (INTCEC 2025) \\ 
15-16 September 2025, Chicago-USA}}
\begin{abstract}
Adversarial robustness remains a critical challenge in deploying neural network classifiers, particularly in real-time systems where ground-truth labels are unavailable during inference. This paper investigates \textit{Volatility in Certainty} (VC), a recently proposed, label-free metric that quantifies irregularities in model confidence by measuring the dispersion of sorted softmax outputs. Specifically, VC is defined as the average squared log-ratio of adjacent certainty values, capturing local fluctuations in model output smoothness. We evaluate VC as a proxy for classification accuracy and as an indicator of adversarial drift. Experiments are conducted on artificial neural networks (ANNs) and convolutional neural networks (CNNs) trained on MNIST, as well as a regularized VGG-like model trained on CIFAR-10. Adversarial examples are generated using the Fast Gradient Sign Method (FGSM) across varying perturbation magnitudes. In addition, mixed test sets are created by gradually introducing adversarial contamination to assess VC's sensitivity under incremental distribution shifts. Our results reveal a strong negative correlation between classification accuracy and $\log(\text{VC})$ (correlation $\rho < -0.90$ in most cases), suggesting that VC effectively reflects performance degradation without requiring labeled data. These findings position VC as a scalable, architecture-agnostic, and real-time performance metric suitable for early-warning systems in safety-critical applications.
\end{abstract}



\section{Introduction}

Deploying artificial intelligence (AI) models on edge devices is increasingly common due to the need for low-latency, privacy-preserving, and resource-efficient inference. These devices, such as embedded processors and microcontrollers, operate in decentralized, bandwidth-limited environments \cite{singh2023edge}. They are critical in mission-driven applications like autonomous navigation \cite{ahmadsmc, vehicular, rezaifsDD}, industrial monitoring \cite{chowdhury2024performance}, and real-time health diagnostics \cite{sarkar2024data, 10022242}. However, limited computational power and the absence of ground-truth supervision at runtime present significant challenges for ensuring reliability and robustness \cite{rezasmc, missionbased, vahidsmc}.

A major issue in edge AI is maintaining generalization when operational data diverge from the training distribution \cite{xu2012robustness}. Since ground-truth labels are unavailable during inference, conventional supervised performance metrics become ineffective \cite{goldwasser2020beyond}. Therefore, autonomous, efficient, and architecture-agnostic performance detection methods are needed. Early detection is particularly valuable, as it enables proactive responses before failures occur.

Model generalization is typically assessed using unseen data, either from the same (IID) or different (OOD) distributions \cite{liu2021towards, gu2021beyond}. While IID validation reflects generalization within the training domain, it does not account for domain shift, sensor drift, or adversarial threats \cite{10316046, zoutowards}.

Current validation methods rely heavily on labels, which are rarely available on edge devices \cite{deng2021labels}. This limitation has led to interest in “label-free” confidence metrics. The logit margin \cite{elsayed2018large}, softmax entropy, and maximum softmax probability (MSP) \cite{hendrycks2017baseline} have emerged as popular tools. High margins and low entropy suggest confident predictions, whereas low margins indicate ambiguity.

Recent work has explored structural indicators of uncertainty, including smoothness and dispersion of softmax outputs \cite{jiangpredicting}, curvature sensitivity \cite{moosavi2017robustness}, local intrinsic dimensionality \cite{ma2018characterizing}, and distributional shift detection \cite{ren2019likelihood}.

In this context, we introduce a new label-free metric, termed \textit{Volatility in Certainty} (VC), which captures local irregularities in softmax-based confidence profiles. The metric reflects the smoothness of the output distribution, with lower VC values indicating more coherent and confident predictions.

We evaluate VC on models trained on MNIST (ANN and CNN) and CIFAR-10 (VGG-like CNN) under adversarial perturbation using the Fast Gradient Sign Method (FGSM) as OOD data generating Methods \cite{oh2022boosting}. Our findings reveal that log-transformed VC shows a strong negative correlation with model accuracy (Pearson $\rho < -0.90$), even under incremental adversarial contamination. To the best of our knowledge, this is the first use of sorted-log-ratio dispersion as a standalone indicator for adversarial robustness and OOD performance estimation without labels.

\section{Proposed Methodology}
\subsection{Objective}

Let $h_w : \mathcal{X} \rightarrow \Delta^C$ be a neural network classifier trained on labeled pairs $(x, y) \in \mathcal{X} \times \mathcal{Y}$, where $\Delta^C$ is the $C$-dimensional probability simplex.

This study explores the \textbf{Volatility in Certainty (VC)} as a label-free metric for evaluating model performance. We focus on two objectives:

\begin{enumerate}
    \item \textbf{VC as a Proxy for Accuracy:}  
We examine the empirical relationship between the VC metric and classification accuracy. Let $\text{Acc}(h_w)$ be the model’s test accuracy and $VC(h_w)$ its volatility score on an unlabeled test set. We aim to show:

\begin{equation}
\text{Corr} \left( \log VC(h_w), \text{Acc}(h_w) \right) \to -1,
\end{equation}

indicating that higher accuracy correlates with smoother confidence transitions (i.e., lower VC). This is tested across various ANN and CNN models trained on MNIST and CIFAR-10.\\

\item \textbf{VC for Adversarial Detection:}  
We assess VC as a signal for adversarial shift. Given a clean test set $\mathcal{D}_{\text{clean}}$, we generate $\mathcal{D}^{(p)}$ by replacing $p\%$ of its samples with FGSM-perturbed images. We then identify the smallest contamination level $p^*$ for which VC shows significant deviation, using a $t$-test:

\begin{equation}
\exists \, p^* : \text{p-value} \left( VC(\mathcal{D}_{\text{clean}}), VC(\mathcal{D}^{(p^*)}) \right) < 0.05.
\end{equation}

This quantifies VC’s potential as an \textbf{early-warning signal} in label-scarce inference environments.
\end{enumerate}

\subsection{Problem Formulation and Metric Definition}

Let $\mathcal{D} = \{x_i\}_{i=1}^N$ be a test set of $N$ images $x_i \in \mathbb{R}^d$. A trained neural network $h: \mathbb{R}^d \rightarrow [0,1]^C$ maps each input to class probabilities:

\begin{equation}
    h(x_i) = \{ P_j(x_i) \}_{j=1}^C, \quad \text{with} \quad \sum_{j=1}^C P_j(x_i) = 1
\end{equation}

Here, $P_j(x_i)$ is the predicted probability for class $j$. Well-trained models tend to produce sharp distributions, one dominant class and others near zero.

In practice, ground-truth labels are often unavailable, complicating validation. To address this, we propose the label-free \textit{Volatility in Certainty (VC)} metric to quantify confidence smoothness. The intuition is that well-generalized models yield consistently sharp predictions, while poor or adversarial models show irregularities.

\textbf{Step 1: Certainty per sample.}  
We define the certainty $\delta_i$ (logit margin) as the difference between the top two predicted probabilities:

\begin{equation}
    \delta_i = \max_j P_j(x_i) - \max_{\substack{j \ne \arg\max_k P_k(x_i)}} P_j(x_i)
    \label{eq:certainty}
\end{equation}

A high $\delta_i$ implies strong confidence; a low value indicates ambiguity. This step is label-independent.

\textbf{Step 2: Sort certainty scores.}  
Form the set $\Delta = \{\delta_i\}_{i=1}^N$ and sort in ascending order:

\begin{equation}
    \delta_1 \le \delta_2 \le \cdots \le \delta_N
\end{equation}

A smooth curve indicates stable predictions; sharp jumps suggest volatility.

\textbf{Step 3: Local volatility.}  
We compute the local volatility $vc_k$ between adjacent certainty values:

\begin{equation}
    vc_k = \left( \log \left( \frac{\delta_{k+1}}{\delta_k + \epsilon_0} \right) \right)^2
    \label{eq:localvolatility}
\end{equation}

Here, $\epsilon_0$ is a small constant (e.g., $10^{-6}$) for numerical stability. This measures relative confidence jumps.

\textbf{Step 4: Compute VC.}  
To reduce noise, we average $vc_k$ over the central 60\% of the sequence:

\begin{equation}
    VC = \frac{1}{0.6N} \sum_{k=0.2N}^{0.8N} vc_k
    \label{eq:VCmetric}
\end{equation}

Lower VC values indicate consistent, confident predictions; higher values suggest instability from adversarial input or poor training.

\begin{algorithm}[H]
\caption{Volatility in Certainty (VC) Computation}
\begin{algorithmic}[1]
\For{$i = 1$ to $N$}
    \State Compute: 
    \[
    \delta_i \gets \max_j P_j(x_i) 
    - \max_{\substack{j \ne \arg\max_j P_j(x_i)}} P_j(x_i)
    \]
\EndFor
\State Sort values: 
\[
\Delta \gets \text{sort}(\{\delta_i\}_{i=1}^N)
\]
\For{$k = 1$ to $N-1$}
    \State $vc_k \gets \left( \log \left( \frac{\delta_{k+1}}{\delta_k + \epsilon_0} \right) \right)^2$
\EndFor
\vspace{0.5em}
\State $VC \gets \dfrac{1}{0.6N} \sum_{k=0.2N}^{0.8N} vc_k$
\vspace{0.5em}
\State \Return $VC$
\end{algorithmic}
\end{algorithm}

\section{Experimental Settings}
This section outlines the datasets, models, and adversarial setups used to evaluate the proposed VC metric.

\subsection{Datasets and Model Architectures}

We validate the VC metric across two vision datasets, MNIST and CIFAR-10, and three neural network architectures of increasing complexity.

\textbf{MNIST} \cite{lecun2002gradientMNIST} is a low-noise, structured dataset of $28 \times 28$ grayscale handwritten digits across ten classes. Its simplicity allows controlled experiments where VC changes can be clearly attributed to perturbations.

\textbf{CIFAR-10} \cite{krizhevsky2009learningCIFAR10} contains $32 \times 32$ RGB images from ten object categories, with higher intra-class variability and natural noise. It offers a more realistic challenge for testing robustness and VC sensitivity.

Using both datasets enables us to evaluate VC in both clean, low-dimensional settings and more complex, real-world-like conditions.

\begin{table*}[t]
\renewcommand{\arraystretch}{1.2}
\caption{Summary of model architectures and training configurations for ANN, CNN, and Regularized VGG models.}
\begin{center}
\label{tab:model-summary}
\begin{tabular}{|l|p{4.2cm}|p{4.2cm}|p{5cm}|}
\hline
\rowcolor[gray]{0.9}
\textbf{Attribute} & \textbf{ANN} & \textbf{CNN} & \textbf{Regularized VGG} \\
\hline
Dataset & MNIST & MNIST & CIFAR-10 \\
\hline
Key Layers & 
FC(784$\rightarrow$128) $\rightarrow$ ReLU $\rightarrow$ FC(128$\rightarrow$64) $\rightarrow$ ReLU $\rightarrow$ FC(64$\rightarrow$10)
&
[Conv(1$\rightarrow$32) $\rightarrow$ ReLU $\rightarrow$ Pool] $\times$1, [Conv(32$\rightarrow$64) $\rightarrow$ ReLU $\rightarrow$ Pool] $\times$1, FC(3136$\rightarrow$128) $\rightarrow$ ReLU $\rightarrow$ FC(128$\rightarrow$10)
&
3 Conv Blocks (64, 128, 256) with BN, ReLU, Pool, Dropout; FC(4096$\rightarrow$512) $\rightarrow$ ReLU $\rightarrow$ FC(512$\rightarrow$10)
\\
\hline
Optimizer & Adam (0.001) & Adam (0.001) & SGD (0.05, momentum=0.9) + StepLR \\
\hline
Label Smoothing & 0.1 & 0.1 & 0.1 \\
\hline
Batch Size & 128 & 128 & 128 \\
\hline
Epochs & 30 & 20 & 100 \\
\hline
Train/Val/Test & 54,000/6,000/10,000 & 54,000/6,000/10,000 & 45,000/5,000/10,000 \\
\hline
\end{tabular}
\end{center}
\end{table*}

We use three models to evaluate VC across architectural complexity:

\textbf{ANN:} A fully connected feedforward model without spatial priors, applied to MNIST. It serves as a baseline to test VC in non-convolutional setups.

\textbf{CNN:} Also trained on MNIST, this model incorporates convolutional layers and spatial inductive bias, enabling better generalization and robustness.

\textbf{Regularized VGG:} For CIFAR-10, we adopt a deeper VGG-inspired network with dropout and weight decay. Its regularization helps resist perturbations, allowing us to test VC's sensitivity in robust models.

This diverse model set ensures VC is tested across architectures with different generalization capabilities.

\subsection{Adversarial Perturbation and VC-Based Detection}

We assess the VC metric under adversarial stress using the Fast Gradient Sign Method (FGSM) \cite{DBLP:journals/corr/GoodfellowSS14}, which perturbs inputs in the direction that maximizes the model’s loss:

\begin{equation}
\tilde{x}_i = x_i + \epsilon \cdot \text{sign} \left( \nabla_x \mathcal{L}(h(x_i), y_i) \right)
\end{equation}

Here, $\epsilon$ is the perturbation magnitude and $\mathcal{L}$ the cross-entropy loss between prediction $h(x_i)$ and true label $y_i$. FGSM is widely used due to its efficiency and effectiveness.

We generate adversarial test sets from the original 10,000 samples using two FGSM schedules:
\begin{itemize}
    \item Fine-grained: $\epsilon \in \{0.000, 0.002, \ldots, 0.030\}$ for sensitivity analysis.
    \item Coarse: $\epsilon = 0.10$ for broader robustness evaluation.
\end{itemize}

At each level, we compute both classification accuracy and $\log(\text{VC})$, then analyze their correlation. In all cases, a strong negative correlation confirms that VC effectively reflects changes in model behavior under adversarial influence.

\section{Results and Observations}
This section presents the corresponding results and key observations.
\subsection{Results of Training Models}
All models were successfully trained with the objective of achieving high classification accuracy and strong generalization performance, as validated on held-out test data. The details of model architecture and training are outlined in Table~\ref{tab:model-summary}. Also, the final training and validation accuracies are summarized in Table~\ref{tab:accuracy-summary}.

\begin{table}[H]
\renewcommand{\arraystretch}{1.4}  
\centering
\caption{Final Training and Validation Accuracy for All Models}
\label{tab:accuracy-summary}
\begin{tabular}{|l|c|c|}
\hline
\rowcolor[gray]{0.9}
\textbf{Model} & \textbf{Train Accuracy} & \textbf{Validation Accuracy} \\
\hline
ANN (MNIST)    & 0.999 & 0.982 \\
\hline
CNN (MNIST)    & 0.999 & 0.994 \\
\hline
VGG (CIFAR-10) & 0.995 & 0.905 \\
\hline
\end{tabular}
\end{table}

As the results indicate, both the ANN and CNN architectures trained on the MNIST dataset exhibit exceptionally high validation accuracy, suggesting strong generalization and minimal overfitting. In contrast, the VGG-based model trained on CIFAR-10, despite its architectural depth and regularization, achieves a relatively lower validation accuracy. This discrepancy reflects the increased visual complexity of CIFAR-10 and potentially greater susceptibility of deeper models to overfitting or adversarial degradation in practice.

\subsection{Correlation Between Log(VC), Accuracy and Sensitivity}
 This section examines the relationship between log(VC) and classification accuracy from three perspectives: first, under two distinct scenarios designed to reveal how these metrics co-vary and indicate model performance in the presence of FGSM perturbations; and second, throughout the training process on clean data.\\

 \subsubsection{\textbf{Scenario A}}

We assess $\log(\text{VC})$ as a surrogate for accuracy across three models: an ANN and CNN on MNIST, and a VGG-style model on CIFAR-10. For each, we create 25 test sets by randomly selecting 1,000 clean images and replacing \( n \in \{0, 5, \ldots, 100\} \) with FGSM-perturbed samples ($\epsilon = 0.10$), introducing controlled adversarial contamination.

\begin{figure}[htp]
    \centering
    \includegraphics[width=0.89\columnwidth]{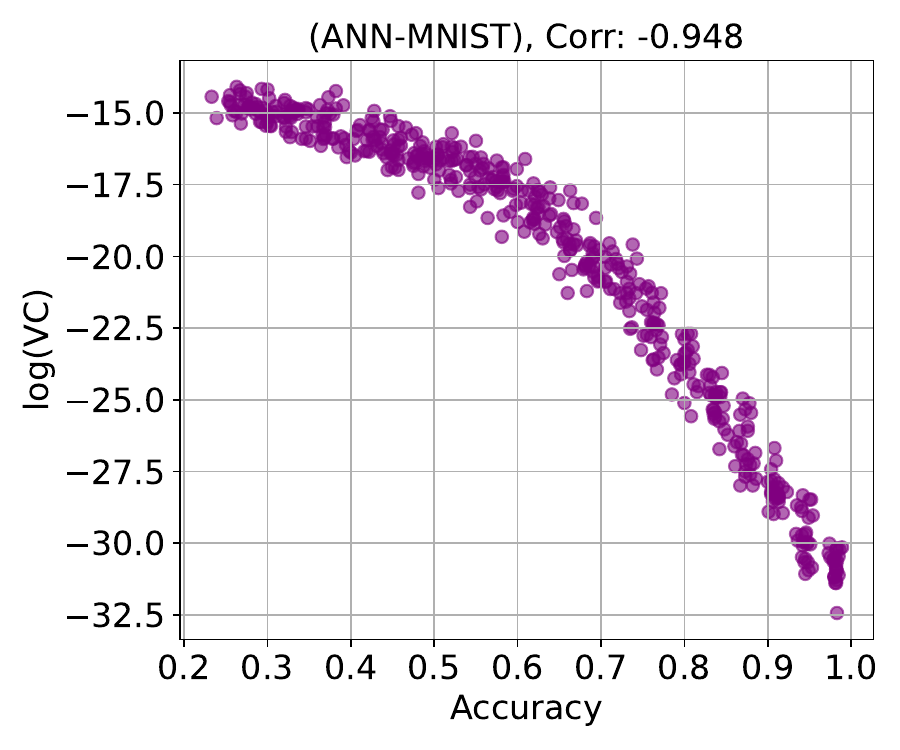}
    \caption{Inverse correlation between model accuracy and $\log(\text{VC})$ for an ANN on MNIST, indicating $\log(\text{VC})$ as a reliable marker of generalization loss.}

    \label{fig:vc_corr_ann}
    \vspace{-1em}
\end{figure}

\begin{figure}[htp]
    \centering
    \includegraphics[width=0.89\columnwidth]{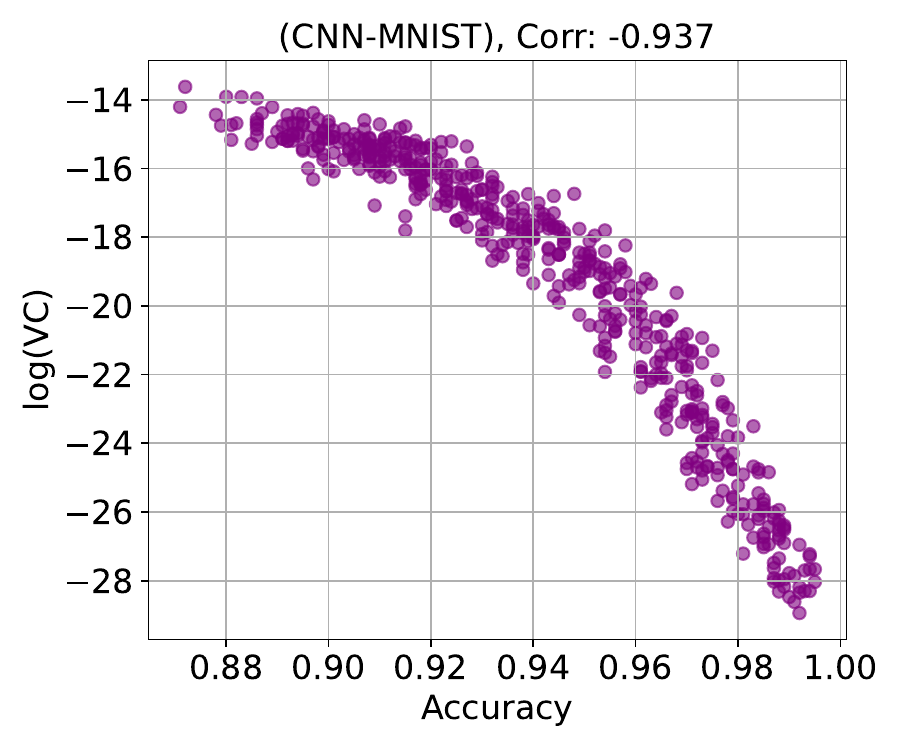}
    \caption{Inverse correlation between model accuracy and $\log(\text{VC})$ for a CNN on MNIST, highlighting $\log(\text{VC})$ as a sensitive generalization indicator.}

    \label{fig:vc_correlation}
    \vspace{-1em}
\end{figure}

\begin{figure}[tp]
    \centering
    \includegraphics[width=0.89\columnwidth]{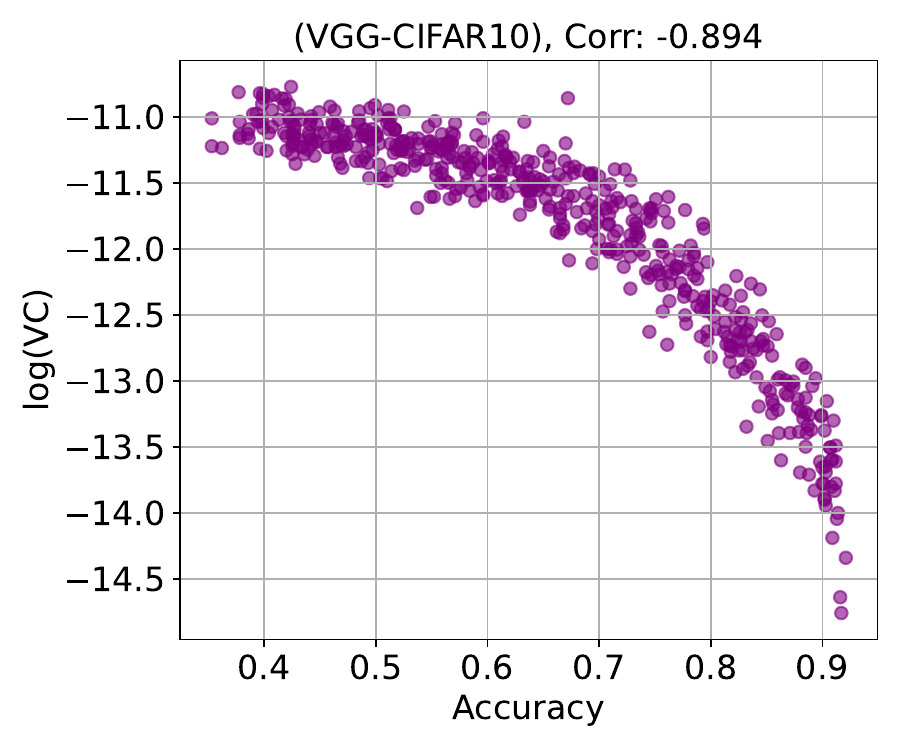}
    \caption{Inverse correlation between model accuracy and $\log(\text{VC})$ for a VGG network on CIFAR-10, confirming its effectiveness as a generalization indicator.}

    \label{fig:VGG_all_trials}
\end{figure}

Figures~\ref{fig:vc_corr_ann}–\ref{fig:VGG_all_trials} plot accuracy vs. $\log(\text{VC})$ across these sets. As adversarial ratio increases, accuracy declines while $\log(\text{VC})$ rises, reflecting greater uncertainty. Strong inverse correlations are observed: $\rho \approx -0.94$ for ANN/CNN (MNIST) and $\rho \approx -0.89$ for VGG (CIFAR-10), which also shows steeper accuracy drops, indicating higher sensitivity.

Overall, $\log(\text{VC})$ consistently tracks performance degradation across models and datasets. It operates without labels, derived solely from softmax outputs, making it ideal for real-time, label-free monitoring on edge and autonomous systems.

We further validate its sensitivity via two-sample $t$-tests between clean and perturbed sets. Even with 5\% contamination, $\log(\text{VC})$ differences are statistically significant (p < 0.05), confirming its responsiveness to distributional shifts.

In summary, $\log(\text{VC})$ is a reliable, unsupervised metric strongly correlated with accuracy and effective for detecting adversarial degradation in label-scarce settings.\\

\subsubsection{\textbf{Scenario B}}

\begin{figure}[htp]
    \centering
    \begin{subfigure}[b]{0.79\linewidth}
        \centering
        \includegraphics[width=\linewidth]{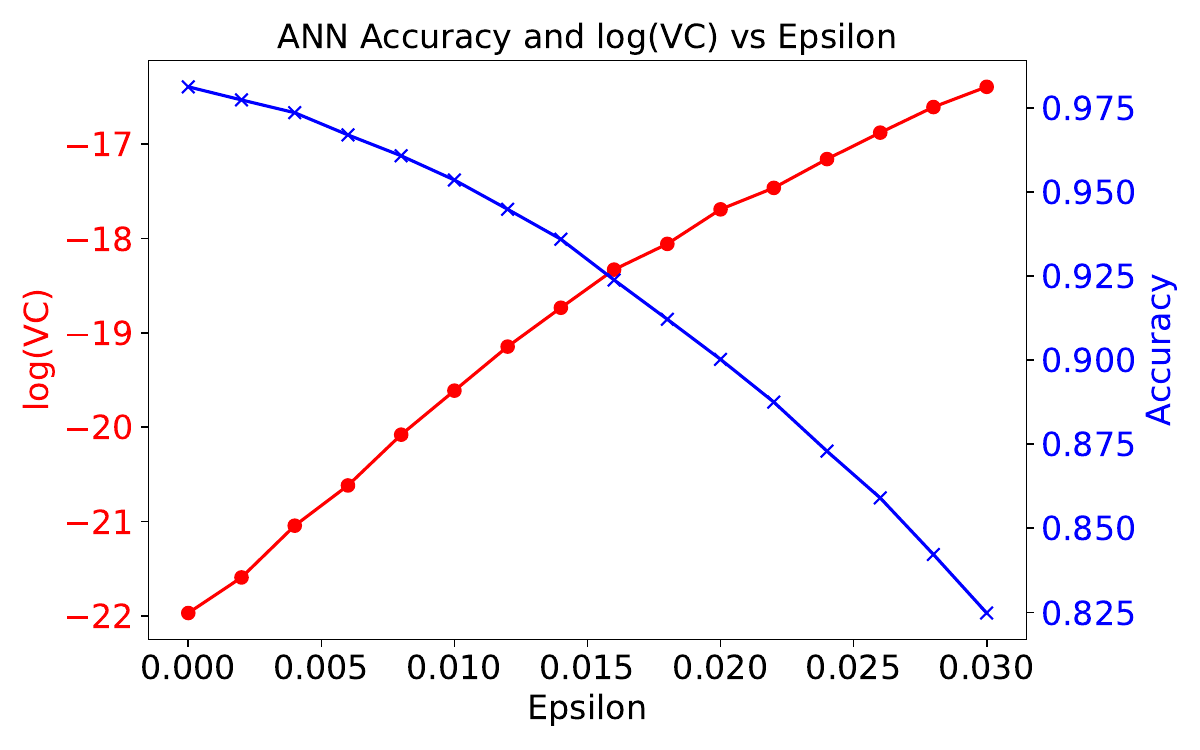}
        \caption{Accuracy vs. $\log(\text{VC})$ for ANN on FGSM-perturbed MNIST with $\epsilon \in [0.000, 0.030]$. Strong inverse correlation observed ($\rho = -0.95$).}
        \label{fig:vc_accuracy_mnist_ann_a}
    \end{subfigure}
    \begin{subfigure}[b]{0.79\linewidth}
        \centering
        \includegraphics[width=\linewidth]{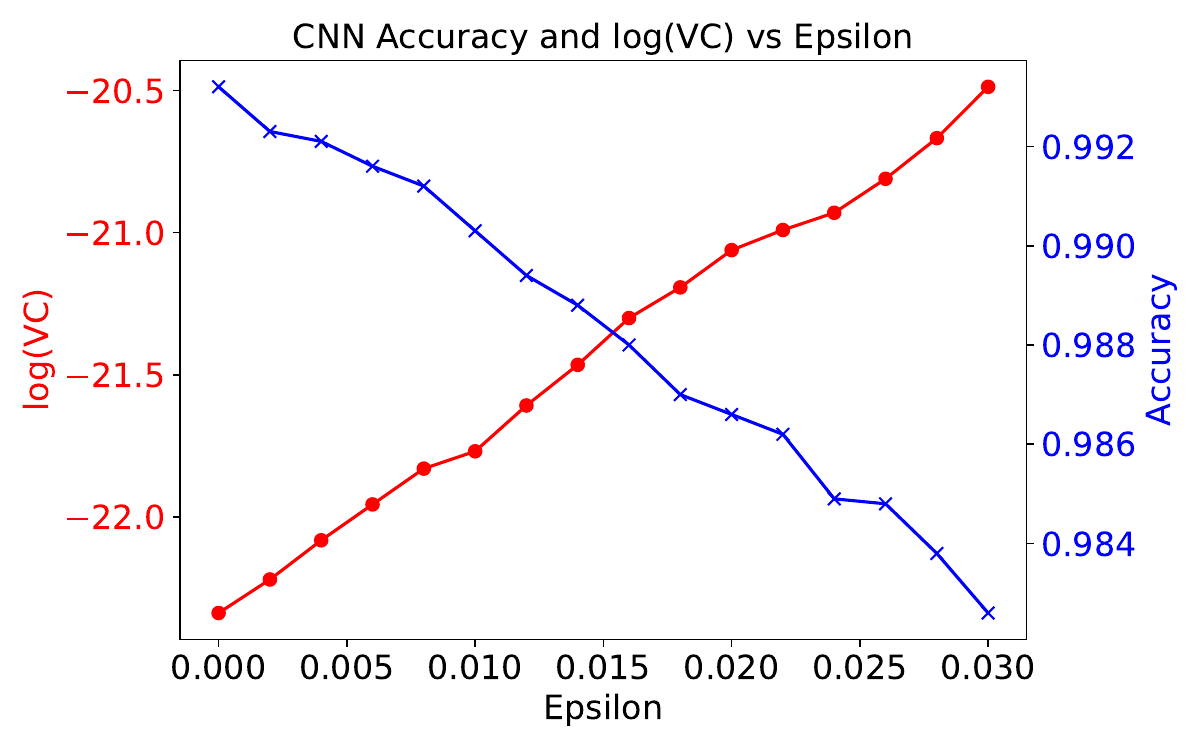}
        \caption{Same for CNN with nearly perfect inverse correlation ($\rho = -0.99$), confirming $\log(\text{VC})$ as a sensitive degradation indicator.}
        \label{fig:vc_accuracy_mnist_ann_b}
    \end{subfigure}
    \caption{Accuracy vs. $\log(\text{VC})$ for ANN and CNN models on FGSM-perturbed MNIST data.}
    \label{fig:vc_accuracy_mnist_ann}
\end{figure}

\begin{figure}[htpb]
    \centering
    \includegraphics[width=0.79\columnwidth]{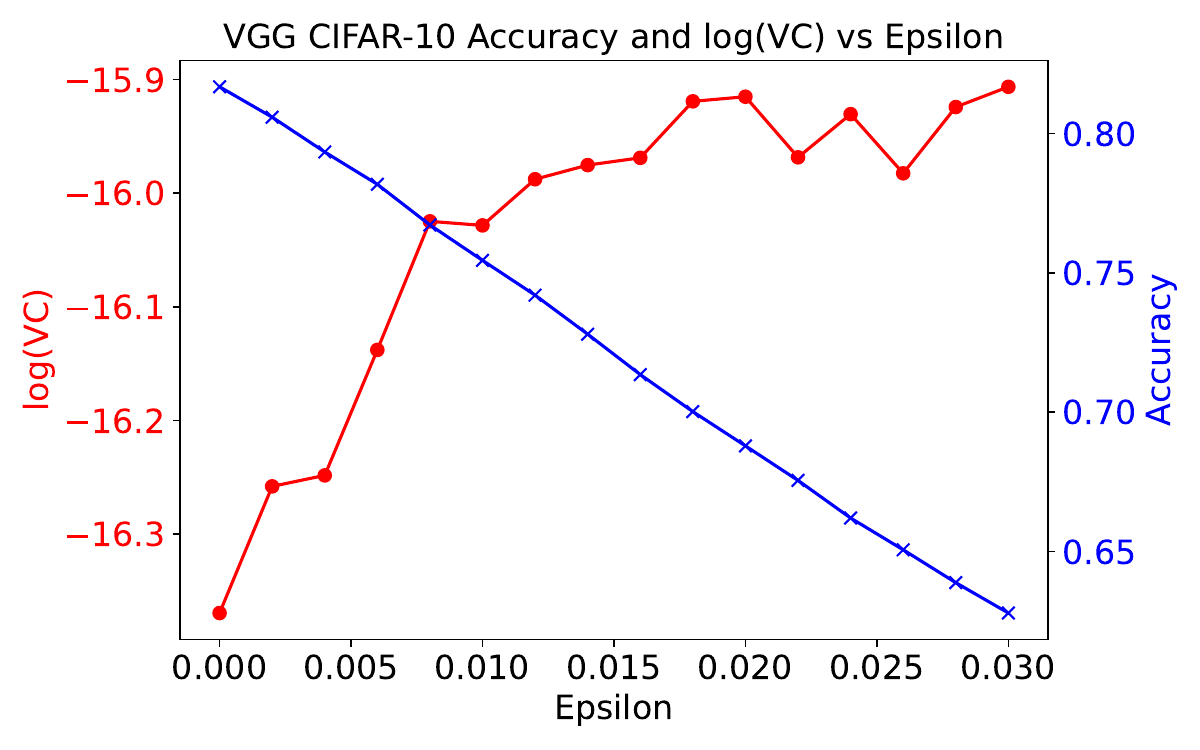}
    \caption{Accuracy vs. $\log(\text{VC})$ for VGG on FGSM-CIFAR-10 ($n=10000$, $\epsilon \in [0.000, 0.030]$). Strong inverse correlation ($\rho = -0.85$) reveals VGG’s greater vulnerability.}

   \label{fig:vc_accuracy_cifar_ann}
\end{figure}

We analyze the correlation between accuracy and $\log(\text{VC})$ under fine-grained FGSM perturbations, this time applying noise to the full test set ($n=10000$). Results are shown in Figures~\ref{fig:vc_accuracy_mnist_ann}–\ref{fig:vc_accuracy_cifar_ann}.

At low perturbation levels ($\epsilon < 0.03$), $\log(\text{VC})$ remains highly correlated with accuracy:
\begin{itemize}
    \item ANN (MNIST):~~~ $\rho = -0.952$
    \item CNN (MNIST):~~~ $\rho = -0.994$
    \item VGG (CIFAR-10): $\rho = -0.850$
\end{itemize}
This inverse correlation is strongest in high-accuracy regimes, where models are well-calibrated and perturbations are initially subtle.

We attribute this to training dynamics: while minimizing loss, models also develop smoother and more stable softmax confidence landscapes. Well-trained models (e.g., ANN and CNN on MNIST) show consistent, structured confidence even on unseen data. In contrast, less generalizable models like VGG on CIFAR-10 display erratic confidence profiles, suggesting instability in the decision surface and greater sensitivity to adversarial noise.\\

\subsubsection{\textbf{Log(VC) and Accuracy Evolve Through Training}}

\begin{figure}[htpb]
    \centering
    \includegraphics[width=0.89\columnwidth]{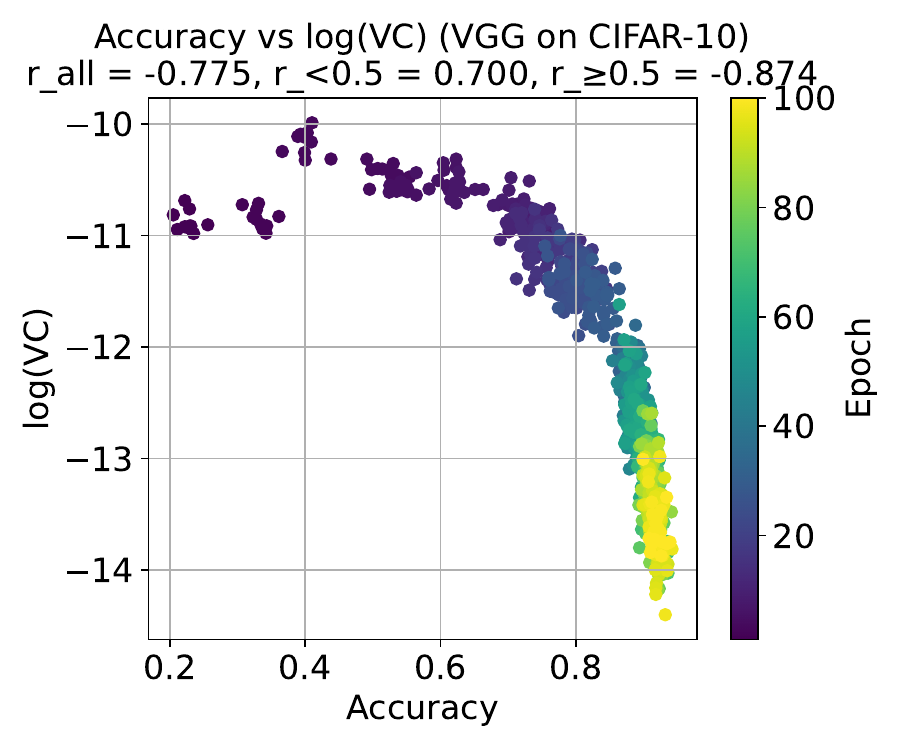}
    \caption{Accuracy vs. $\log(\text{VC})$ during training. Each point (epoch) is evaluated on 10 validation sets. Color shows training progress. The inverse trend highlights growing confidence and convergence.}
    \label{fig:VGG_Train}
\end{figure}

To examine how the relationship between accuracy and $\log(\text{VC})$ develops during training, we tracked both metrics across epochs using only clean data. At each epoch, accuracy and VC were evaluated on ten random validation sets of 1,000 images to ensure robust trend estimation. Color in Figure~\ref{fig:VGG_Train} encodes training progression, visualizing the temporal dynamics.

As training advances and accuracy improves, a strong inverse correlation emerges ($\rho = -0.878$ for Accuracy $> 0.5$). Early in training, however, when accuracy is low and predictions unstable, the relationship is weak ($\rho = -0.132$ for Accuracy $\le 0.5$). This suggests that $\log(\text{VC})$ becomes more reliable as model confidence stabilizes.

Figure~\ref{fig:VGG_Train} can be generated during training to estimate accuracy trends in real time. In settings without labels, $\log(\text{VC})$ offers a practical approximation of model performance.

While this interpretation is theoretical, it aligns with observed patterns in softmax distributions. Further study could clarify how softmax confidence evolves and affects VC sensitivity.

In summary, $\log(\text{VC})$ is most effective when the model is well-trained and confident. It provides a sensitive, label-free signal for performance monitoring and early detection of reliability loss in real-time systems.





\section{Conclusion}
This paper introduces Volatility in Certainty (VC), a novel label-free metric for detecting adversarial perturbations in neural network classifiers. By quantifying local irregularities in sorted softmax outputs, VC correlates strongly with classification accuracy and sensitively identifies distributional shifts without access to ground-truth labels. Our experiments on MNIST and CIFAR-10 across multiple model architectures demonstrate the practical utility of VC in real-time, label-scarce scenarios. We have shown that the VC metric is effective for benchmarking classifier performance in the absence of labels. In future work, we aim to integrate the VC metric into training pipelines to encourage models that generalize better to out-of-distribution (OOD) scenarios. By incorporating VC-based regularization or early stopping criteria, it may be possible to guide training toward smoother and more robust decision boundaries. Furthermore, we aim to investigate the use of VC as a proactive indicator for OOD generalization performance, enabling early estimation of deployment-time robustness without requiring labeled validation data.
 
\bibliography{Refer}
\bibliographystyle{unsrt}

\end{document}